\documentclass[runningheads]{llncs}
\usepackage{graphicx}
\usepackage{latexsym}
\usepackage{listings}

\usepackage{soul}
\usepackage{algorithm}
\usepackage{algpseudocode}
\usepackage[british]{babel}

\begin{document}
\title{Clash of the Explainers: \\ Argumentation for Context-Appropriate Explanations}

\titlerunning{Argumentation for Context-Appropriate Explanations}
%

\author{Leila Methnani \inst{1}\orcidID{0000-0002-9808-2037} \& Virginia Dignum \inst{1}\orcidID{0000-0001-7409-5813}\&  Andreas Theodorou \inst{1}\orcidID{0000-0001-9499-1535}}

\author{Leila Methnani\inst{1}\orcidID{0000-0002-9808-2037} \and
Virginia Dignum\inst{1}\orcidID{0000-0001-7409-5813} \and
Andreas Theodorou\inst{1}\orcidID{0000-0001-9499-1535}}
\authorrunning{L. Methnani et al.}
%
\institute{Umeå University, Sweden \\
\email{leila.methnani@umu.se, virginia@cs.umu.se, \\ andreas.theodorou@umu.se}}

\authorrunning{L. Methnani et al.}
%

\maketitle              
\begin{abstract}

Understanding when and why to apply any given eXplainable Artificial Intelligence (XAI) technique is not a straightforward task. There is no single approach that is best suited for a given context. This paper aims to address the challenge of selecting the most appropriate explainer given the context in which an explanation is required. For AI explainability to be effective, explanations and how they are presented needs to be oriented towards the stakeholder receiving the explanation. If---in general---no single explanation technique surpasses the rest, then reasoning over the available methods is required in order to select one that is context-appropriate. Due to the transparency they afford, we propose employing argumentation techniques to reach an agreement over the most suitable explainers from a given set of possible explainers. 

In this paper, we propose a modular reasoning system consisting of a given mental model of the relevant stakeholder, a reasoner component that solves the argumentation problem generated by a multi-explainer component, and an AI model that is to be explained suitably to the stakeholder of interest. By formalising supporting premises---and inferences---we can map stakeholder characteristics to those of explanation techniques. This allows us to reason over the techniques and prioritise the best one for the given context, while also offering transparency into the selection decision.

\keywords{Explainability  \and Transparency \and Argumentation.}
\end{abstract}

\bibliographystyle{splncs04}

\section{Introduction} \label{Introduction}

Now that the need for eXplainable Artificial Intelligence (XAI) has been firmly established \cite{arrieta2020explainable}, the development of state-of-the-art techniques, such as Local Interpretable Model-agnostic Explainations (LIME) \cite{ribeiro2016should} and SHapley Additive exPlanations (SHAP) \cite{lundberg2017unified}, is continuously being pursued. While researchers and practitioners have benefited from model interpretations offered by such techniques, issues remain that make them tricky to adopt. One identified pitfall of current XAI methods is the failure to make their limitations clear; misleading explanations can be inconspicuous, and may result in downstream actions that are unjustified \cite{XAIforPractitioners,kaur2020interpreting,slack2020fooling}. Moreover, with this large suite of XAI methods comes the need to understand each in order to make an informed choice about which to select, and then further interpret the results received. The assumption here is that these explainees will have the expertise to make such informed choices and analyses. In reality, current explanations mostly serve system developers---who have limited temporal constraints (i.e. how fast the explanation is to be produced and can be consumed) and require a large amount of in-depth information \cite{miller2019explanation}. 

Thus, we propose an XAI system built to support the selection of an XAI method using symbolic reasoning, taking into account what the explainee may \textit{value} and \textit{need} in order to determine the optimal method and explanations to present. In this work, we focus only on selecting the best suited explainer. While an aggregated view of many potential explanations could perhaps be more appropriate depending on context, it is possible to mislead the stakeholder when attempting to augment one explanation using other explanations generated by incompatible techniques \cite{XAIforPractitioners}. 
 
Moreover, it is important to consider the context within which explanations are needed and the context within which they will be used. When we refer to \textit{context sensitivity}, we mean capturing the frame that surrounds the request for an explanation. This includes capturing relevant knowledge about available explanation techniques, such as strengths and weaknesses, as well as relevant knowledge about the target audience, referred to as the \textit{stakeholder}, including their understanding of the AI system and their intentions behind seeking an explanation for its decision-making. Too few techniques are developed with the intention of modelling the stakeholder's view of the system and with consideration for their explainability needs \cite{bhatt2020machine}. Still, explainability begins with considering \textit{who} is in need of an explanation. Often, the `who' is not entirely aware of what type of explanation is best suited to their needs, which introduces the added challenge of selecting an optimal explanation technique that can be sufficiently interpreted for the context at hand. Thus, more transparency, i.e. providing insights into `why' an explanation has been selection, is required.

\paragraph{Contribution}
Our main contribution is a framework for the formalisation of and reasoning over: i)the characteristics of explanation techniques; ii) the properties that make them well-suited---or ill-suited---towards various contexts; and iii) the contexts in which the produced explanations are to be used. With such a formalisation, transparency into the selection of existing explainer methods that also maps needs to their capabilities is afforded. This is proposed by our introduction of the reasoning component on top of a multi-explainer system. The intention is to take an appropriate mental model of a target explainee into account when offering explanations, removing any assumption that the user is sufficiently informed of all the strengths and weaknesses of each explanation method and the technicalities required to interpret the output. For the purposes of this paper only, we assume that these explainee mental models are given and accurately portray their existing knowledge and intentions.

In this paper, we begin with an overview in Section~\ref{Background} of the concepts that underpin our proposed solution; we define terms central to explainable artificial intelligence (XAI) and human-centric XAI (HCXAI), as well as computational reasoning and argumentation for both making and justifying decisions. In Section~\ref{MxRS}, we describe our working example and describe an experimental setup. 
Finally in Section~\ref{Conclusion}, we conclude that reasoning over multiple explanations can be used to map explainers to explainee characteristics. We further motivate the need for evaluating this work in a human-subject study to validate the transparency effects with target human stakeholders. 

\section{Background} \label{Background}

\subsection{Explainable Artificial Intelligence}
It is widely agreed that no one `true' explanation exists and that stakeholders have diverse needs when it comes to explainability and interpretability of AI models \cite{arya2019one}. There can be many explanations for a single outcome, each contributing towards a particular explanatory dimension \cite{XAIforPractitioners}. 

To build on this notion of \textit{explanatory dimensions} we first need to unpack what is meant by an \textit{explanation}. Here, we adopt the definition offered by Guidotti \textit{et al.} In their survey, they describe an explanation to be `an interface' between humans and a decision maker that is ... both an accurate proxy of the decision maker and comprehensible to humans'' \cite{guidotti2018survey}. Markus \textit{et al.} characterise explainability similarly by highlighting the properties of \textit{interpretability}---relating to clarity and parsimony---and \textit{fidelity}---relating to completeness and soundness \cite{markus2020role}. The importance of each property, they argue, is dependent on the reason that explainability is demanded in the first place. Today, a vast amount of XAI methods have been---and continue to be---developed towards different explanatory demands. Models can be developed to be \textit{intrinsically} explanatory and interpretable to humans. Consider decision trees, for instance, as such intrinsically interpretable Machine Learning (ML) models; decision-making steps can be intuitively followed and understood for simple trees\footnote{It is worth noting that as decision trees scale, their interpretability may also decline due to the sheer size of the structure.}. When such interpretable models suffer in terms of predictive power, the need for more complex models may arise, motivating the employment of \textit{black-box} models that are not intuitively understandable. \textit{Post-hoc} XAI methods become useful in this case, where explanations can be offered after model training has been completed. More useful still, are \textit{model-agnostic} methods, which are those that are not innately baked into a specific ML model and can---in principle---be applied to any ML model. The benefits of taking a model-agnostic approach include the freedom to choose any machine learning model for the prediction task at hand and \textit{still} being able to offer an explanation to stakeholders after the fact. It offers practitioners the added benefit of comparing interpretability across machine learning methods as well, which can be insightful in terms trade-offs. These qualities make post-hoc, model-agnostic XAI approaches desirable, and we therefore focus on these methods only.

Post-hoc methods can be \textit{global} or \textit{local}. If an understanding of the overall behaviour of an AI system is required, then global explanation techniques can be employed. For an understanding of how a single decision came to be, local techniques can be used. In this paper, we consider both classes of XAI techniques as the context within which explanations are needed will drive the choice of class.

\subsection{Human-centred Explainable Artificial Intelligence}
Effectively applying XAI techniques requires, according to Pitman and Munn, practitioners to ``start from the consumer's place of understanding and build upon it'' \cite{XAIforPractitioners}. This is precisely what Human-centred Explainable Artificial Intelligence (HCXAI) intends to do by putting the human at the core of XAI design \cite{ehsan2020human}.

HCXAI moves beyond superficial consideration of who the target human might be; by employing human-computer interaction (HCI) techniques such as \textit{value-sensitive design} (VSD) \cite{ehn2017scandinavian} and \textit{participatory design} \cite{friedman2013value}, HCXAI aims to involve the human stakeholder directly for a holistic understanding of relevant design requirements. VSD is rooted in consideration for human values when designing technology. Participatory design aims to build technology together with those who hold a stake in it. These approaches often result in the formulation of \textit{mental models} of relevant human stakeholders.

Here, we define a mental model to be the stakeholder's cognitive representation of how any given system works. This includes what the stakeholder knows about the system's components, the interactions between them, as well as the processes that transform them \cite{carroll1988mental}. Additionally, we consider what is known about the stakeholder themselves; this includes their \textit{needs} to fulfil a role and adhere to values, amongst others. Mental models are important to consider in the context of explainable AI in particular for numerous reasons. For one, there is no one-size-fits-all solution to explanations---or to transparency at large---suited for all stakeholders and their needs \cite{Theodorou2017ConnectionScience}. Furthermore, individuals within stakeholder groups may also have slightly different requirements based on their personal experience with AI in particular. Similarly, as we humans explain concepts amongst ourselves, we tailor our explanations based on the explainee and what they are already expected to understand about the concept, building on top of that expected knowledge. Moreover, at different times there might be different needs. Attention is given to context, which we understand as the relevant elements that shape the setting within which an explanation is required. This calls for the consideration of relevant explainee characteristics mapped to XAI techniques to inform more suitable selections of explanations.

\subsection{Reasoning and Argumentation} 
Transparency into the various elements that contribute towards the choice of an explanation technique offers stakeholders the value of making more informed decisions when it comes to AI and its applications in industry. Transparency is afforded through the neat and intuitive way of forming \textit{inferences}---drawing conclusions---derived from a knowledge base of \textit{facts} already assumed to be true. The availability of such a knowledge base of truths together with well-formulated and documented rules of inference can help stakeholders in their interpretation of any given explanation.

Argumentation is considered pivotal to the way humans arrive at conclusions and thus make decisions
\cite{dietz2022argumentation}. Arguments are formulated to defend and persuade given claims and further support actions that can be taken. Offering ``the correct'' argument is not of the essence, which is fitting in the context of offering explanations---we do not necessarily have one \textit{correct} approach to explaining a decision outcome, but rather many; some approaches are more suitable than others depending on the context. 

Computational argumentation can therefore offer a solid foundation for human-centric AI, where the intention is to augment human cognition. Thus, we choose to adopt a computational argumentation approach to reasoning over the choice of explanation when given a system of multiple explainers suited for presenting different motivations for the outcome of a single machine learning model. 

Consider Dung's Abstract Argumentation (AA) framework \cite{dung1995acceptability}. Formally, the AA framework takes the form of a pair $ S = \langle \mathit{Ar, R} \rangle $ where $\mathit{Ar}$ is the set of arguments, and $\mathit{R}$ is a binary relation of \textit{attacks}, where 
$\mathit{R} \subseteq Ar \times Ar$. For $\mathit{a}, \mathit{b} \in \mathit{Ar}$, $\mathit{R(a, b)}$ indicates that the argument $\mathit{a}$ represents an attack on the argument $\mathit{b}$. 

To compute a reasonable \textit{position} given by the argumentation framework $S$, an \textit{extension} set $E$ can be built, where $E \subseteq Ar$. $E$ is considered \textit{conflict-free} if no member within it attacks another. That is there exists no elements a and b within $E$ such that $R(a,b)$. An argument $\mathit{a}$ that is attacked by $\mathit{b}$ through the relation $\mathit{R(a, b)}$ is said to be defended by $\mathit{c}$ if $\mathit{R(c, a)}$ and thus \textit{acceptable} as an argument with respect to the position. 
Conflict-free positions that contain acceptable arguments are said to be \textit{admissible}. \cite{thimm2014strategic}

Label-based approaches to solving arguments also exist \cite{baroni2011introduction}. Arguments that are free from the effect of any attack are labelled \textit{IN}. IN arguments are accepted and render any argument they attack to be \textit{OUT}. Therefore, we can reject all arguments that are labelled OUT. Arguments that are neither IN nor OUT are undecided and labelled \textit{UNDEC}. This label-based approach can be depicted visually using a graphical representation of the argument framework. The AA framework can be constructed as a graph; the nodes represent arguments and directed edges represent attack relations. Starting with a graph of the abstract argument, the label-based approach can be used to remove those nodes labelled as OUT, together with any outgoing edges from OUT nodes, and retain those nodes that are labelled as IN. What is left is a neat visual depiction of an accepted position as shown in Figure~\ref{fig:solver}. Visualisations of the AA framework as a graph support the value of transparency, offering a representation of the argumentation flow that is often easily digestible for humans.

\section{Related Work}

In his work summarising XAI insights from the social sciences, Miller \cite{miller2019explanation} describes some requirements for the selection of the ``best'' explanation in AI. From a social sciences perspective, the question asked by the explainee is of utmost importance, and often driven by anomalies or surprising observations. As users interact with system, they learn and generalise, adjusting their need for certain explanations along the way. This motivates the need for selection mechanisms that take into account both the question asked by the explainee, and the mental model they have of the system. Further motivation is given for maintaining a model of both the explainee and the explainer, with early work in XAI such as Cawsey's EDGE system \cite{cawsey1993planning} or Weiner's BLAH system \cite{weiner1980blah} that both promote explanations that are oriented towards the users and what they seem to \textit{know} about the system. The focus, however, is consistently on selecting and evaluating individual explanations rather than motivating the existing methods that generate them, which is also of significant importance. 

Considering the time and expertise currently required to select an XAI method that generates relevant explanations, motivation exists to automate the whole XAI pipeline, as in Automated Machine Learning (AutoML). The objective with AutoML is to automate the ML pipeline end to end as a means of enabling domain experts to create ML solutions without much of the technical pre-requisites \cite{he2021automl}. In their work, Cugny \textit{et al.} \cite{cugny2022autoxai} propose a framework for AutoXAI, also motivated by the need for contextualising XAI solutions and relieving the data scientist of the tedious tasks required to do so. There are three main components in their framework: (1) the user who offers elements of the context, (2) the context adapter that selects a subset of explainers that match these specified needs, and (3) a hyper parameter optimiser that performs a search over hyperparameters to reduce loss based on explanation evaluation function aggregates. As with most XAI techniques, this framework is also oriented towards the practitioner---namely data scientists---as its primary user. In our work, we motivate consideration for a wider scope that includes expert and non-expert users alike. Moreover, the authors raise ethical issues that may follow from the explanation selection, namely the bias that may arise from preference configurations within their framework. Biases that may arise from stakeholder preferences is one consideration that encourages our choice of argumentation and reasoning; they support making the facts, opinions, and beliefs that drive such preferences explicit and clear. 

The Gorgias Argumentation Framework is a structured argumentation framework that accounts for beliefs and how they shape the conditions that lead to a given position. Gorgias allows for the generation of priority arguments, where preference or relative strength between arguments can be expressed. These assigned strengths then determine their attack or defense influence over other arguments. In their work, Kakas \textit{et al.} explore Gorgias output in relation to XAI and the need for ``socially useful'' explanations, arguing that the properties such as being (1) attributive in the rules of an argument and (2) contrastive in the set of preferences presented are supported by argumentation frameworks such as Gorgias \cite{spanoudakis2022gorgias}. In their work developing Visual Gorgias, Vassiliades \textit{et al.} \cite{vassiliades2021visual} describe how an added visualisation layer that offers graphical representations can support the user's understanding of the argumentation framework and the decisions that were made.

\section{Modular Multi-explainer and Reasoning System} \label{MxRS}
In this section, we describe our proposed solution---a modular explanation selection system for determining context-appropriate explanations. We consider contextual factors such as the stakeholder receiving the explanation, timeliness of the explanation, the model that needs to be explained, the application domain, etc. A high-level overview of the system is offered by the illustration in Figure~\ref{fig:reasoning_system}. One critical component is a given mental model of the relevant stakeholder. Here, we will assume that the mental model is already defined and provided. It captures a representation of the stakeholder's cognitive state with respect to the context including elements such as values and requirements. This representation contributes towards a Knowledge Base (KB) of \textit{facts} and \textit{beliefs} that the reasoner component can access to make inferences over an appropriate explanation offered by multiple \textit{explainers}. We consider multiple explanation techniques to comprise of the multi-explainer component of this system. Each explainer has characteristics that support particular values, requirements, and other knowledge characteristics of the users that employ them. Thus, we can extend our KB with facts about these explainers, and construct relations in the form of \textit{attacks} as captured by Dung's AA framework. Our final component is ML system itself, which can---in principle---be any dataset and model of the stakeholder's choice. In our system, we promote the selection of post-hoc and model-agnostic explainers for the multi-explainer component, thus allowing for the desirable flexibility of choosing any ML model to explain. End-to-end, we make available any supporting premises---and inferences---that map stakeholder characteristics to those of the explanation techniques to make the process not only transparent, but also \textit{contestable}. The latter refers to the property of a system to provide information why this was the best decision possible \cite{aler2020contestable}. In our argumentation framework, we demonstrate that we have selected the best possible explanation---given the conditions presented to the system---through our attacks system. 

We see a system like this being utilised when companies are in need of a formal and standardised approach to fulfilling the explainability demands of their stakeholders based on regulatory and other policy requirements. While a system like this is by no means intended to replace the interview and participatory design processes that an organisation is expected to engage in with their stakeholders, it offers a means of concretising those findings and making any biases explicit. In the context of this paper, the decision problem we are concerned with is that of selecting an explanation technique when an explanation is demanded. Capturing and presenting various assumptions about the context to the stakeholder supports their understanding of organisational decision making around AI. It can also support organisations in their due diligence whenever their AI ecosystem is audited. Presenting reasoning around explainability becomes increasingly important when you consider the fact that AI explanations can be `weaponised.' It is possible, for example, to induce over-confidence in an AI system by generating explanation that are misleading \cite{slack2020fooling,molnar2020interpretable,ehsan2022human}. We propose our framework as a means of designing for accountability and avoiding ``ethics washing.''

\subsection{Use Case: Predicting Current Housing Prices}
To best explain our work, we offer an example use case and describe a modular Multi-explainer and Reasoning System (MxRS) component by component. Consider the scenario of buying and selling houses. A real-estate agency has deployed an AI system trained on historical housing data of the region. The agency is performing evaluations for a customer interesting in selling their house. This customer is also seeking an explanation for the price point offered by the prediction system. The agency has an in-house team of AI architects who design, develop, and maintain their AI ecosystem together with a suite of XAI techniques. Now, the agency must make a decision on how to explain the outcome to the customer; a choice that is contextual and stakeholder dependent. For the purposes of our use case, we define an instance of a system using our proposed architecture component by component. We use this theoretical implementation to ground our discussion throughout.

\subsubsection{The Stakeholder Mental Model(s) and Building Context}
Collecting the knowledge required to construct a mental model can be achieved through many means; for instance, using sensors, direct user input, or a mix of the two. The HCXAI community advocates for conversational explainable AI, where users can engage in conversation when seeking explanations from a given AI system \cite{miller2019explanation,lakkaraju2022rethinking}. Neurosymbolic techniques can therefore be employed to extract symbolic rules from the user's natural language prompts if such a dialogue were to be developed as the interface between our proposed system and the user. The extracted symbolic knowledge can then be used by the MxRS to construct a mental model for the stakeholders and populate the knowledge base for the system to reason over an appropriate explanation technique to employ. 

The appropriate solution will depend on (at least) both stakeholders presented in our use case. Facts and beliefs about each side should be considered for the context at hand. Let us take the first stakeholder to be the real-estate agency that has deployed the housing price prediction model and the second to be the customer interested in selling their house; the former must present the latter with an explanation as to why the selling price is not as high as they had expected.
 
Consider that the real-estate agency has two XAI techniques at their disposal when explaining their model outcomes. The first one is LIME, a local interpretability technique that is considered to be human-friendly and intuitive to decipher, the second is counterfactual, describing how the variables need to be adjusted to obtain a different outcome. For this scenario, we assume that both these methods are valid. The real estate agency expresses a preference for computationally cheap methods over those that generate short explanations, which are considered more human-friendly. They also express a preference for using more trustworthy techniques over those that generate short explanation due to the observation that their customers have expressed doubt towards their services; they want to boost the customer's trust. They want to demonstrate to their trustworthiness to the customer by being transparent about their methods; those that are susceptible to adversarial attack is a stronger argument against trustworthiness than instability.

To build the argumentation framework around this decision problem of selecting a context-appropriate explainer, we must first identify the knowledge related to this selection process. In the housing prices scenario, we consider the requirements from the two stakeholders, the real estate agency and the customer, which include:

\begin{itemize}
    \item The real estate agency has preference for computationally cheap methods. 
    \item The customer puts high values on the trustworthiness of an explanation.
    \item The customer requires a `human-friendly' explanation.    
\end{itemize}

Let us also list some contextual facts and some beliefs about the context in general:

\begin{itemize}
    \item It has been shown that LIME is susceptible to adversarial attack that can intentionally mislead explainees by hiding biases \cite{slack2020fooling}.
    \item An explanation is human-friendly' if it is short (presenting only one or two causes) and contrastive, i.e., it compares the current context with some context in which the event would not have happened. \cite{molnar2020interpretable,miller2019explanation}
    \item Simpler explanations that boost the likelihood of the explainee understanding and accepting an explanation may better support trust than offering a more likely explanation. \cite{miller2019explanation}
\end{itemize}

\subsubsection{Contextual Information about the ML System}
In our MxRS, the ML component is more than just an ML Model and its output; multiple explainers would in fact require access to the dataset, as depicted by Figure~\ref{fig:reasoning_system}. Explainers may, for example, generate alternative outputs based on permutation or in order to select samples as example-based explanations. So we define our ML system component to consist of metadata, data, and the trained ML model. Metadata capture information about the dataset that may be relevant to the context and used to populate the KB. This can include information extracted from a datasheet, describing the dataset \cite{gebru2021datasheets}, or a model card reporting on the trained model \cite{mitchell2019model}.

In our use case, the dataset is historical housing data of the region within which the stakeholder is looking to buy and sell. Making predictions for house prices is considered a regression problem solved by models such as \textit{eXtreme Gradient Boosting} (XGBoost) \cite{bell2022s}. XGBoost is a tree-based ensemble learning method that has high predicting power for regression problems.

\subsubsection{The Multi-Explainer Component}
The multi-explainer component consists of at least two \textit{explainer} sub-components. Each explainer provides its own means of extracting or producing explanations from the ML model.  For example, one explainer may be providing local explanations by using the Local Interpretable Model-agnostic Explanations (LIME) \cite{ribeiro2016should} method and another may offer counterfactual explanations using Diverse Counterfactual Example (DiCE) \cite{mothilal2020explaining}.

In principle, a multi-explainer component can comprise multiple instances of the same explanation technique, only with different parameter settings. Then, the explanations offered will be intrinsic and/or model-specific, therefore requiring modifications and substitutions within the ML System module of our architecture. Such an approach is outside the scope of this paper but relevant to consider for future iterations.

Each explainer also contains a list of arguments for why its produced explanations are the best for each explanation request. It offers those arguments to the reasoner component. We would also like to highlight the ability to take into consideration various characteristic of explanation techniques themselves within explainer arguments. Such characteristics of interest may include the computational costs of generating an explanation using any given method, the environmental impact of said computation, or even access to interpretable visualisations for example. These pros and cons may be mapped to stakeholder requirements and contribute to the argumentation computation.

\subsubsection{The Reasoner Component}\label{reasoner component}

The reasoner component is made up of a KB and an argumentation solver. The KB is populated with relevant facts and beliefs offered by the previously described multi-explainer component, as well as those offered by the mental model of the stakeholder, to be described in the upcoming subsection. The argumentation solver computes admissible positions given the arguments posed. Our working example uses notation from both Dung's Abstract Argumentation Framework \cite{dung1995acceptability} and Gorgias Preference-based Argumentation Framework \cite{kakas2020abduction}; introducing preferences offers a value-based approach that is appropriate in various applications of our systems due to emphasis on human-centricity when solving for context-appropriate explanations. If the context requires consistency and involves arguments constructed from imperfect information, then admissibility-based semantics are desirable \cite{baroni2011introduction}. In principle, however, an instance of our proposed system can be implemented with any argumentation semantics; while the outcome of the system will be determined by it, transparency will illuminate biases in the designer's choice. 

Using syntax from the Gorgias framework, we can represent knowledge of rules, conflicts, and preferences using predicate symbols. Labelled rules are constructed using the form $rule(Label, Head, Body)$. Here, the Head is a literal, the Body is a list of literals, and the Label is a compound term made up of the rule's name along with selected variables from the Head and Body \cite{noel2009gorgias}. To represent negative literals, Gorgias uses the form $neg(L)$. An attack is characterised both by the complements and preferences defined in the framework. That is, argument $a$ is said to attack argument $b$ if they have complementary conclusions, and argument $a$ contains rules of higher or equal priority to argument $b$. To follow, we put the scenario described in our use case above into Gorgias syntax to construct the argumentation framework and solve for the queries neg(use(X=lime)) and use(X=counterfactual) respectively.

\begin{lstlisting}[breaklines=true]

%% Arguments, where X is an explainer
rule(r1(X), use(X), [is_sparse(X)]).
rule(r2(X), neg(use(X)), [neg(is_computationally_cheap(X))]).
rule(r3(X), use(X),[is_trustworthy(X)]).
rule(r4(X), is_trustworthy(X), [is_stable(X)].
rule(r5(X), neg(is_trustworthy(X)), [susceptible_to_adversarial_attack(X)].

%% Preference Rules
rule(pr1(X), prefer(r2(X), r1(X)), []). % prefer computational costs short explanations.
rule(pr2(X), prefer(r3(X), r2(X)), []). % prefer trustworthiness over  computational costs.
rule(pr3(X), prefer(r5(X), r4(X)), []). % susceptibility to adversarial attack is stronger argument than stability.

%% Facts / Beliefs
rule(f1, is_sparse(X = counterfactual), []).
rule(f2, is_sparse(X = lime), []).
rule(f3, is_computationally_cheap(X = lime), []).
rule(f4, neg(is_computationally_cheap(X = counterfactual)), []).
rule(f5, susceptible_to_adversarial_attack(X = lime), []).
rule(f6, neg(is_stable(X = lime)), []).

\end{lstlisting}

The set of arguments, presented as rules $r1$ through to $r5$, together with the preference rules pr1 through to pr3, already encapsulate bias in the designer's choice and presenting these rules and reasoning steps to any stakeholder (customers, auditors, system engineers, etc.) will make the designer's assumptions clear. Starting with $r1$, we see that the agency has a rule to use methods that produce short explanations, which is characterised by \textit{sparsity}. If it is not computationally cheap, $r2$ says not to use the method. Using trustworthy methods is depicted by $r3$. What constitutes a trustworthy method is determined by the \textit{stability} of the method, as described by $r4$, while $r5$ states that a method's trustworthiness is compromised if it is susceptible to adversarial attack. We can also see that the agency has a preference to save on computational costs over generating short explanations, as described by $pr1$. More important still is the preference for trustworthiness; $pr2$ says it is a stronger argument over computational cost. With regards to trustworthiness, $pr3$ states that susceptibility to adversarial attack is a stronger argument against it than stability is for it. Then, we can populate the knowledge base with facts and/or beliefs about the available explainers, for instance $f5$, that states LIME is susceptible to adversarial attack.

\begin{figure}
    \centering
    \includegraphics[width=9cm,keepaspectratio]{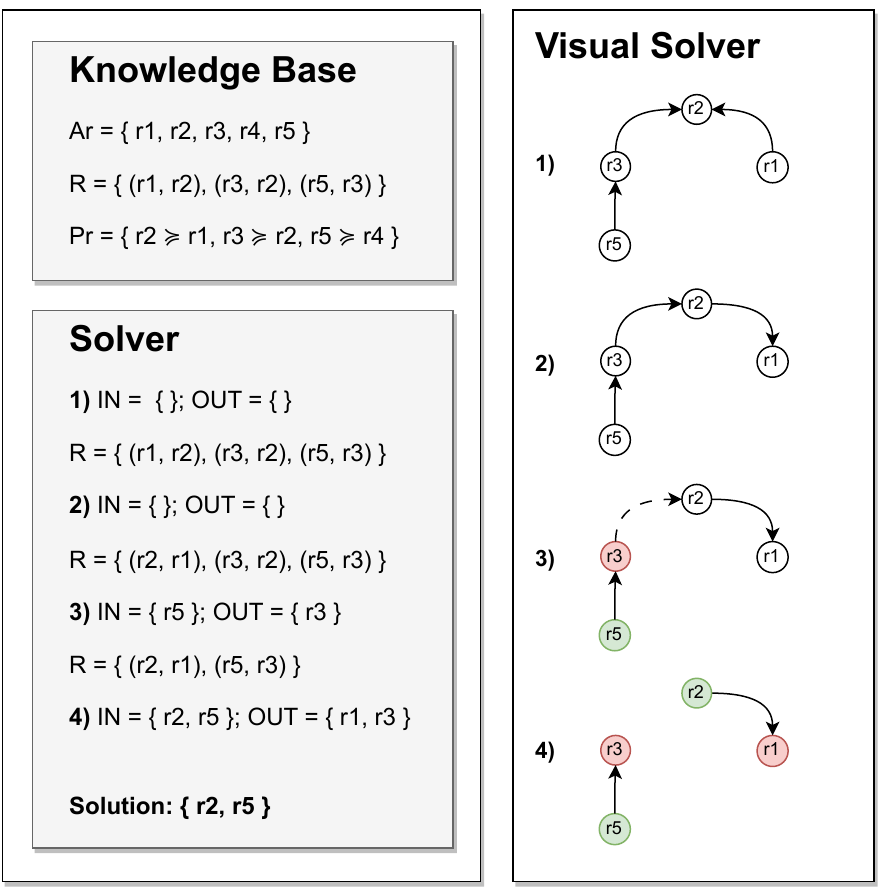}
    \caption{An example solution using labellings, where the argument not to use LIME holds in the described argumentation framework. The example is further described in Sub-section~\ref{reasoner component}.}
    \label{fig:solver}
\end{figure}

In Figure~\ref{fig:solver}, the KB consists of an argument set $Ar$, binary relations $R$ representing attacks, and preferences $Pr$ that depict priority of some arguments over others. The example depicted in the figure solves for the argument against using LIME as an explainer in the described context; i.e. the framework is queried to see if r2(X=lime): neg(use(X=lime)) holds. The arguments and relations generated using this query are $Ar = \{ r1,r2,r3,r5 \}$ and $R = \{ (r1,r2), (r3,r2), (r5,r3) \}$. The argument $r4$ is not generated because it does not hold that LIME is stable. The preferences are $Pr = \{ r2 \succeq r1, r3 \succeq r2, r5 \succeq r4\}$. Graphically, we can represent these arguments as nodes and the attacks as directed edges. Priority arguments are handled by countering an incident attack. So, in step 2, we can flip the edge in the graph and update the attack relations to apply this priority preference. Recall from section~2 that an argument without any effective attacks is considered IN (that is, any attack is labelled OUT). Thus, in step 3, the argument $r5$ can be trivially marked as IN (coloured green in Figure~\ref{fig:solver}) and any argument that $r5$ attacks will be marked as OUT (coloured red in Figure~\ref{fig:solver}). Uncoloured nodes are UNDEC. We can disregard attacks coming from OUT arguments, as indicated by the dotted edges in Figure~\ref{fig:solver}. Removing those outgoing edges in step 4 allows us to consider $r2$ an argument that is IN---it considered \textit{defended} by $r5$. It follows that any argument attacked by $r2$ is OUT. Finally, we can see that only two arguments remain, concluding an admissible and accepted position $\{ r2,r5 \}$, which contains the argument r2(X=lime): neg(use(X=lime)), that is, not to use LIME in the given context, holds.

\begin{figure}
    \centering
    \includegraphics[width=9cm,keepaspectratio]{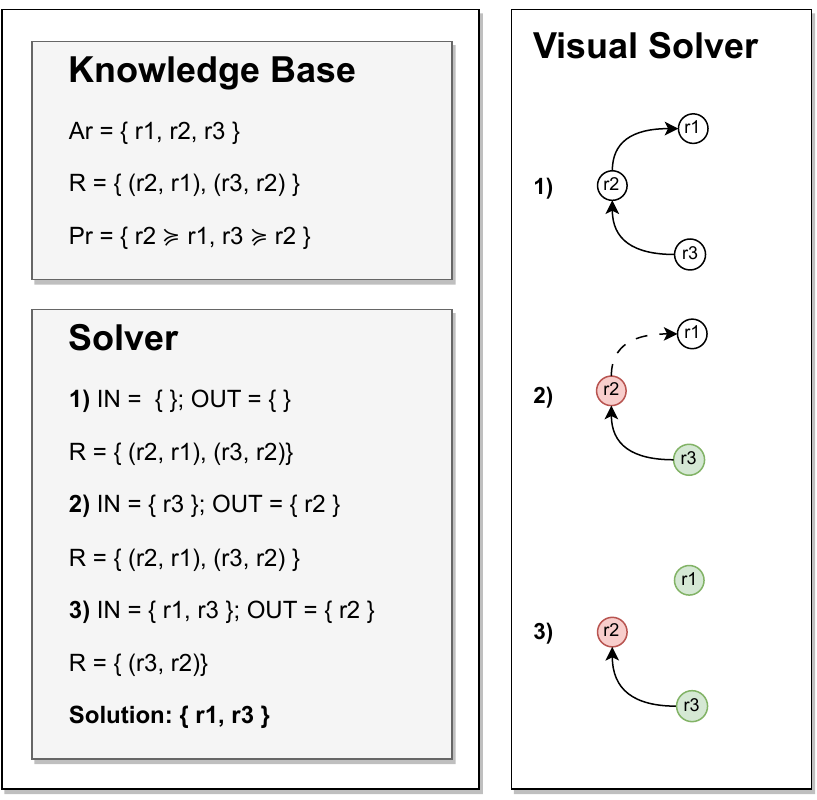}
    \caption{An example solution using laballeings, where the argument to use counterfactual explanations holds in the described argumentation framework. The example is further described in Sub-section~\ref{reasoner component}.}
    \label{fig:solver2}
\end{figure}

In Figure~\ref{fig:solver2}, we demonstrate the same process, but with the query use(X=counterfactual), showing that it also holds. The agency may therefore select a counterfactual method to generate explanations for the explainee in this given context. The agency may apply DiCE, for example, and present the customer with a diverse set of feature-perturbed instances of their house that would have received the price they were expecting. In natural language, such an explanation would be interpreted as follows, ``you would have received the price range you are looking for if your house had an additional balcony.'' Such an explanation shows how the number of balconies influences housing prices in the given geographic location, for example, and can support the explainee in understanding what renovations may---or may not---support them in achieving their target price. This explanation is more context-appropriate considering the explainee's question of why their expectations on housing prices were not met by the agency's evaluation; LIME may have offered an explanation that depicts the number of balconies as an influential factor on price, but would not have offered the additional information required to satisfy the explainee's question about a specific target price bracket.   

It may be that an empty set $E = \emptyset$ is returned by the solver. In such a case, we propose that the system always present a default explanation as a fallback. Here, we are of the position that some explanation is better than none. 
We prioritise both ensuring an explanation is always available regardless of the context, and that the system is transparent at all times, even when only a trivial solution exists. The opportunity to offer a log showing the reasoning steps that result in  $E = \emptyset$ also exists here, allowing users to better understand where conflicts in arguments arise and how that might impact their understanding of the system as a whole. 

\begin{figure}
    \centering
    \includegraphics[width=12cm,keepaspectratio]{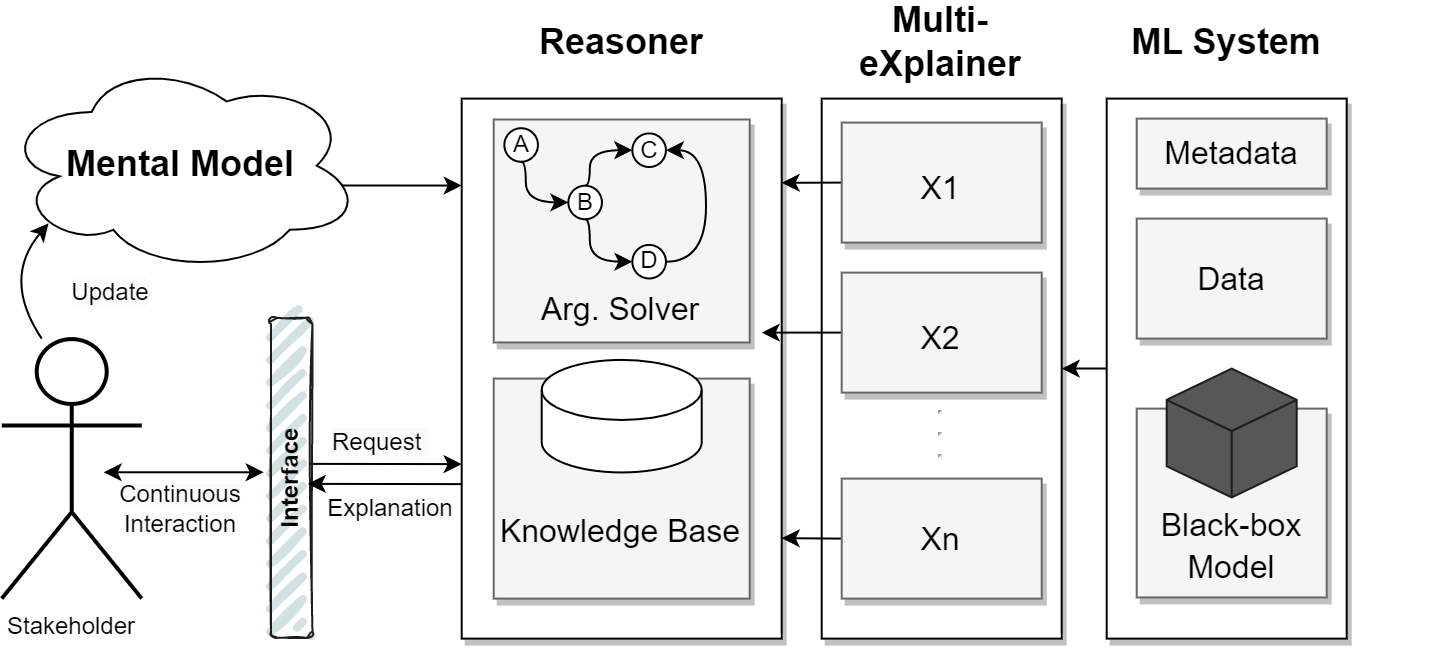}
    \caption{A high-level illustration of our proposed explanation selection system comprising of: (a) the mental model; (b) the Reasoner component that consists of a Argumentation Solver and a Knowledge Base; (c) the multi-explainer component consisting of many explainers; and (d) the Machine Learning system consisting of any available meta-data, a dataset, and the trained model itself that is to be explained (black-box).}
    \label{fig:reasoning_system}
\end{figure}

Notice that the stakeholder's perception of the system will not only be influenced by the explanations presented, but also by the transparency by which an explanation is selected. By offering the stakeholder insights through the provision of all reasoning steps for explanation selection, the system is also influencing their understanding of an explanation context-dependency, thereby influencing the state of the stakeholder's mental model. Therefore, interaction with both the ML model and our explanation selection system as a whole will result in the need for continuous updates to the mental model of the stakeholder, as depicted in our high-level diagram in Figure~\ref{fig:reasoning_system}. At this stage, considerations for how such updates can be done have not been made, but doing so through interactivity and accounting for user feedback is one way we propose this work to move forward.

\section{Conclusion and Future Work} \label{Conclusion}
In selecting a suitable explainer, there is no one-size-fits-all solution. There are increasingly many methods to select from when it comes to explaining AI outcomes. Some are more appropriate than others given the context, which includes the AI system(s) being used, the mental model of the explainee and the questions they ask, and potentially also legal or sectorial requirements on what are suitable explanations. Still, the path to selecting context-appropriate solutions is not always conflict-free considering the facts and beliefs that give shape to the context. Using argumentation to reason over available explanation techniques and select that which will generate a context-appropriate explanation is therefore desirable. Moreover, making those reasoning steps accessible and readily available for the target stakeholder offers transparency into explanation selection, making XAI applications themselves less of an opaque practice.

Beyond selecting explanations, evaluating the quality of the selection and the explanation requires grounding in studies with human subjects. Therefore, we propose the development of a Minimum Viable Product (MVP) and a user study as the necessary next steps to determining the effectiveness of our approach in collaboration with target stakeholders.
Future work includes such an MVP implementation, along with an investigation of how neurosymbolic AI techniques can be utilised for extracting additional contextual knowledge and beliefs as a supplement to traditional methods of participatory design for constructing stakeholder mental models.
\appendix

\section*{Ethical Statement}

The authors have no competing interests, research involving human participants and/or animals, or issues of informed consent to disclose.

\bibliography{ecai}

\begin{thebibliography}{10}
\providecommand{\url}[1]{\texttt{#1}}
\providecommand{\urlprefix}{URL }
\providecommand{\doi}[1]{https://doi.org/#1}

\bibitem{aler2020contestable}
Aler~Tubella, A., Theodorou, A., Dignum, V., Michael, L.: Contestable black
  boxes. In: Rules and Reasoning: 4th International Joint Conference, RuleML+
  RR 2020, Oslo, Norway, June 29--July 1, 2020, Proceedings 4. pp. 159--167.
  Springer (2020)

\bibitem{arrieta2020explainable}
Arrieta, A.B., D{\'\i}az-Rodr{\'\i}guez, N., Del~Ser, J., Bennetot, A., Tabik,
  S., Barbado, A., Garc{\'\i}a, S., Gil-L{\'o}pez, S., Molina, D., Benjamins,
  R., et~al.: Explainable artificial intelligence (xai): Concepts, taxonomies,
  opportunities and challenges toward responsible ai. Information fusion
  \textbf{58},  82--115 (2020)

\bibitem{arya2019one}
Arya, V., Bellamy, R.K., Chen, P.Y., Dhurandhar, A., Hind, M., Hoffman, S.C.,
  Houde, S., Liao, Q.V., Luss, R., Mojsilovi{\'c}, A., et~al.: One explanation
  does not fit all: A toolkit and taxonomy of ai explainability techniques.
  arXiv preprint arXiv:1909.03012  (2019)

\bibitem{baroni2011introduction}
Baroni, P., Caminada, M., Giacomin, M.: An introduction to argumentation
  semantics. The knowledge engineering review  \textbf{26}(4),  365--410 (2011)

\bibitem{bell2022s}
Bell, A., Solano-Kamaiko, I., Nov, O., Stoyanovich, J.: It’s just not that
  simple: an empirical study of the accuracy-explainability trade-off in
  machine learning for public policy. In: 2022 ACM Conference on Fairness,
  Accountability, and Transparency. pp. 248--266 (2022)

\bibitem{bhatt2020machine}
Bhatt, U., Andrus, M., Weller, A., Xiang, A.: Machine learning explainability
  for external stakeholders. arXiv preprint arXiv:2007.05408  (2020)

\bibitem{carroll1988mental}
Carroll, J.M., Olson, J.R.: Mental models in human-computer interaction.
  Handbook of human-computer interaction pp. 45--65 (1988)

\bibitem{cawsey1993planning}
Cawsey, A.: Planning interactive explanations. International Journal of
  Man-Machine Studies  \textbf{38}(2),  169--199 (1993)

\bibitem{cugny2022autoxai}
Cugny, R., Aligon, J., Chevalier, M., Roman~Jimenez, G., Teste, O.: Autoxai: A
  framework to automatically select the most adapted xai solution. In:
  Proceedings of the 31st ACM International Conference on Information \&
  Knowledge Management. pp. 315--324 (2022)

\bibitem{dietz2022argumentation}
Dietz, E., Kakas, A., Michael, L.: Argumentation: A calculus for human-centric
  ai. Frontiers in Artificial Intelligence  \textbf{5} (2022)

\bibitem{dung1995acceptability}
Dung, P.M.: On the acceptability of arguments and its fundamental role in
  nonmonotonic reasoning, logic programming and n-person games. Artificial
  intelligence  \textbf{77}(2),  321--357 (1995)

\bibitem{ehn2017scandinavian}
Ehn, P.: Scandinavian design: On participation and skill. In: Participatory
  design, pp. 41--77. CRC Press (2017)

\bibitem{ehsan2020human}
Ehsan, U., Riedl, M.O.: Human-centered explainable ai: Towards a reflective
  sociotechnical approach. In: International Conference on Human-Computer
  Interaction. pp. 449--466. Springer (2020)

\bibitem{ehsan2022human}
Ehsan, U., Wintersberger, P., Liao, Q.V., Watkins, E.A., Manger, C.,
  Daum{\'e}~III, H., Riener, A., Riedl, M.O.: Human-centered explainable ai
  (hcxai): beyond opening the black-box of ai. In: CHI Conference on Human
  Factors in Computing Systems Extended Abstracts. pp.~1--7 (2022)

\bibitem{friedman2013value}
Friedman, B., Kahn, P.H., Borning, A., Huldtgren, A.: Value sensitive design
  and information systems. Early engagement and new technologies: Opening up
  the laboratory pp. 55--95 (2013)

\bibitem{gebru2021datasheets}
Gebru, T., Morgenstern, J., Vecchione, B., Vaughan, J.W., Wallach, H., Iii,
  H.D., Crawford, K.: Datasheets for datasets. Communications of the ACM
  \textbf{64}(12),  86--92 (2021)

\bibitem{guidotti2018survey}
Guidotti, R., Monreale, A., Ruggieri, S., Turini, F., Giannotti, F., Pedreschi,
  D.: A survey of methods for explaining black box models. ACM computing
  surveys (CSUR)  \textbf{51}(5),  1--42 (2018)

\bibitem{he2021automl}
He, X., Zhao, K., Chu, X.: Automl: A survey of the state-of-the-art.
  Knowledge-Based Systems  \textbf{212},  106622 (2021)

\bibitem{kakas2020abduction}
Kakas, A., Michael, L.: Abduction and argumentation for explainable machine
  learning: a position survey. arXiv preprint arXiv:2010.12896  (2020)

\bibitem{kaur2020interpreting}
Kaur, H., Nori, H., Jenkins, S., Caruana, R., Wallach, H., Wortman~Vaughan, J.:
  Interpreting interpretability: understanding data scientists' use of
  interpretability tools for machine learning. In: Proceedings of the 2020 CHI
  conference on human factors in computing systems. pp. 1--14 (2020)

\bibitem{lakkaraju2022rethinking}
Lakkaraju, H., Slack, D., Chen, Y., Tan, C., Singh, S.: Rethinking
  explainability as a dialogue: A practitioner's perspective. arXiv preprint
  arXiv:2202.01875  (2022)

\bibitem{lundberg2017unified}
Lundberg, S.M., Lee, S.I.: A unified approach to interpreting model
  predictions. Advances in neural information processing systems  \textbf{30}
  (2017)

\bibitem{markus2020role}
Markus, A.F., Kors, J.A., Rijnbeek, P.R.: The role of explainability in
  creating trustworthy artificial intelligence for health care: A comprehensive
  survey of the terminology, design choices, and evaluation strategies. Journal
  of Biomedical Informatics  \textbf{113},  103655 (2021).
  \doi{https://doi.org/10.1016/j.jbi.2020.103655},
  \url{https://www.sciencedirect.com/science/article/pii/S1532046420302835}

\bibitem{miller2019explanation}
Miller, T.: Explanation in artificial intelligence: Insights from the social
  sciences. Artificial intelligence  \textbf{267},  1--38 (2019)

\bibitem{mitchell2019model}
Mitchell, M., Wu, S., Zaldivar, A., Barnes, P., Vasserman, L., Hutchinson, B.,
  Spitzer, E., Raji, I.D., Gebru, T.: Model cards for model reporting. In:
  Proceedings of the conference on fairness, accountability, and transparency.
  pp. 220--229 (2019)

\bibitem{molnar2020interpretable}
Molnar, C.: Interpretable machine learning. Lulu. com (2020)

\bibitem{mothilal2020explaining}
Mothilal, R.K., Sharma, A., Tan, C.: Explaining machine learning classifiers
  through diverse counterfactual explanations. In: Proceedings of the 2020
  conference on fairness, accountability, and transparency. pp. 607--617 (2020)

\bibitem{XAIforPractitioners}
{Munn}, M., {Pitman}, D.: Explainable AI for Practitioners. O'Reilly Media,
  Inc., California (2022)

\bibitem{noel2009gorgias}
No{\"e}l, V., Kakas, A.: Gorgias-c: Extending argumentation with constraint
  solving. In: Logic Programming and Nonmonotonic Reasoning: 10th International
  Conference, LPNMR 2009, Potsdam, Germany, September 14-18, 2009. Proceedings
  10. pp. 535--541. Springer (2009)

\bibitem{ribeiro2016should}
Ribeiro, M.T., Singh, S., Guestrin, C.: " why should i trust you?" explaining
  the predictions of any classifier. In: Proceedings of the 22nd ACM SIGKDD
  international conference on knowledge discovery and data mining. pp.
  1135--1144 (2016)

\bibitem{slack2020fooling}
Slack, D., Hilgard, S., Jia, E., Singh, S., Lakkaraju, H.: Fooling lime and
  shap: Adversarial attacks on post hoc explanation methods. In: Proceedings of
  the AAAI/ACM Conference on AI, Ethics, and Society. pp. 180--186 (2020)

\bibitem{spanoudakis2022gorgias}
Spanoudakis, N.I., Gligoris, G., Kakas, A.C., Koumi, A.: Gorgias cloud: On-line
  explainable argumentation. In: System demonstration at the 9th International
  Conference on Computational Models of Argument (COMMA 2022) (2022)

\bibitem{Theodorou2017ConnectionScience}
Theodorou, A., Wortham, R.H., Bryson, J.J.: Designing and implementing
  transparency for real time inspection of autonomous robots. Connection
  Science  \textbf{29}(3),  230--241 (2017).
  \doi{10.1080/09540091.2017.1310182}

\bibitem{thimm2014strategic}
Thimm, M.: Strategic argumentation in multi-agent systems. KI-K{\"u}nstliche
  Intelligenz  \textbf{28}(3),  159--168 (2014)

\bibitem{vassiliades2021visual}
Vassiliades, A., Papadimitriou, I., Bassiliades, N., Patkos, T.: Visual
  gorgias: A mechanism for the visualization of an argumentation dialogue. In:
  25th Pan-Hellenic Conference on Informatics. pp. 149--154 (2021)

\bibitem{weiner1980blah}
Weiner, J.: Blah, a system which explains its reasoning. Artificial
  intelligence  \textbf{15}(1-2),  19--48 (1980)

\end{thebibliography}

\end{document}